\documentclass[conference]{IEEEtran}
\IEEEoverridecommandlockouts
\usepackage{amsmath,amssymb,amsfonts}
\usepackage{hyperref}
\usepackage{algorithmic}
\usepackage{graphicx}
\usepackage{textcomp}
\usepackage{xcolor}
\usepackage{biblatex}
\usepackage{authblk}
\def\BibTeX{{\rm B\kern-.05em{\sc i\kern-.025em b}\kern-.08em
    T\kern-.1667em\lower.7ex\hbox{E}\kern-.125emX}}
\begin{document}

\title{Research on Travel Route Planing Problem\\Based on Greedy Algorithm}

\author[]{Yiquan Wang}

\affil[]{\textit{College of Mathematics and System Science, Xinjiang University, Urumqi, China}}
\affil[]{\textit{Shenzhen X-institute, Shenzhen, China}}

\affil[]{\texttt{ethan@stu.xju.edu.cn}} 

\maketitle

\begin{abstract}
The route planning problem based on the greedy algorithm represents a method of identifying the optimal or near-optimal route between a given start point and end point. In this paper, the PCA method is employed initially to downscale the city evaluation indexes, extract the key principal components, and then downscale the data using the KMO and TOPSIS algorithms, all of which are based on the MindSpore framework. Secondly, for the dataset that does not pass the KMO test, the entropy weight method and TOPSIS method will be employed for comprehensive evaluation. Finally, a route planning algorithm is proposed and optimised based on the greedy algorithm, which provides personalised route customisation according to the different needs of tourists. In addition, the local travelling efficiency, the time required to visit tourist attractions and the necessary daily breaks are considered in order to reduce the cost and avoid falling into the locally optimal solution.
\end{abstract}


\section{Introduction}

With the rapid development of China's tourism industry and the continuous strengthening of international exchanges, more and more foreign tourists choose to come to China to experience the rich natural scenery and profound cultural heritage. According to the data of the China Migration Administration, the number of foreigners entering China at all ports increased significantly in 2024, especially the number of tourists coming to China reached 4.361 million. In order to better serve these foreign tourists, China has launched a 144-hour transit visa-free policy and implemented it in multiple cities and ports. This policy not only provides more convenience for foreign tourists to travel to China, but also promotes exchanges and cooperation between Chinese and foreign personnel.

In order to enhance the travel experience of foreign tourists, we conducted a study on the attraction routing problem based on a greedy algorithm. This project aims to provide scientific and reasonable tourism route planning for foreign tourists through mathematical modeling and algorithm optimization, ensuring that they can visit as many high-quality attractions as possible within a limited time, while optimizing travel costs and time arrangements.

\section{Related Work}

Nowadays, people are paying more and more attention to personal experience and self-needs during the tourism process \cite{ref1}. Therefore, the use of mathematical modeling methods to automatically generate travel route plans that meet the needs of tourists and help them enjoy a better travel experience is increasingly receiving attention from both academia and industry.

Tourism route planning is one of the popular research issues in the field of smart tourism. The current research work combines the work in operations research with the travel route planning problem in social media, and plans to study it as a variant based on directed off-road or traveling salesman problems \cite{ref2, ref3, ref4, ref5}.

In 2016, Feiran et al. proposed a flexible multi-task deep travel route planning framework called MDTRP, which integrates rich auxiliary information to achieve more effective planning \cite{ref6}. Xinyi et al. proposed an interactive visual analysis method and introduced automatic route optimization algorithms and various interactions to help users optimize and adjust their itineraries for better tourism route planning \cite{ref7}.

Ana et al. evaluated the theoretical basis of multidimensional concepts of tourist spatiotemporal behavior in 2017 and proposed a conceptual model based on behavioral perspectives and intercity destination backgrounds as a comprehensive analytical framework \cite{ref8}. Yuan et al. developed complex statistical models using high-resolution and fine-grained spatiotemporal data in 2019, and established multinomial logit (MNL) models to identify factors that affect tourist destination choices \cite{ref9}. Nithyasri et al. proposed a tourism itinerary generator in 2024. By utilizing user-provided details such as destination, budget, and interests, the system can create personalized travel plans \cite{ref10}.

\section{Methods}

In order to streamline the analytical process and concentrate on the fundamental aspects, we employed the principal component analysis (PCA) technique utilising the MindSpore framework. This approach effectively reduced the dimensionality of the urban evaluation indicators and extracted multiple pivotal principal components. In the case of some datasets that did not pass the KMO test, an innovative approach was taken whereby the entropy weight method was combined with the TOPSIS method in order to achieve a comprehensive evaluation and weight allocation of the data. When constructing the route optimisation model, the greedy algorithm strategy was adopted in order to ensure the optimisation and efficiency of the overall tourism planning scheme.

Our data comes from the Fifth ``Huashu Cup'' National University Student Mathematical Modeling Competition in 2024 \cite{ref11}. It contains 352 cities with 100 attractions in each city, and the information of each attraction contains the name of the attraction, address, description of the attraction, opening hours, and so on.

We adopt a comprehensive evaluation system to simplify the process, and classify the city according to multiple dimensions, such as its size, environmental status, cultural heritage, transportation conditions, as well as local climate and culinary characteristics. Specifically, for the indicators that meet the KMO test criteria, we apply principal component analysis to realize their dimensionality reduction; for the indicators that fail to meet this criterion, we introduce the entropy-based TOPSIS method to complete the simplification to one dimension.

In response to the data collected, where direct modeling resulted in a large number of indicators, this study chose to simplify the dataset by means of dimensionality reduction, which in turn led to the construction of an assessment model for analysis. The KMO test was used as a measure of the effectiveness of the different dimensionality reduction methods, which is mainly used to determine whether the correlation between the variables is sufficient to perform factor analysis. The KMO value ranges from 0 to 1. Generally speaking, when the KMO value exceeds 0.6, the dataset is considered to be suitable for factor analysis. The KMO algorithm can be found in Pseudocode 1.

\begin{figure}[h]
	\centering
	\includegraphics[width=0.48\textwidth]{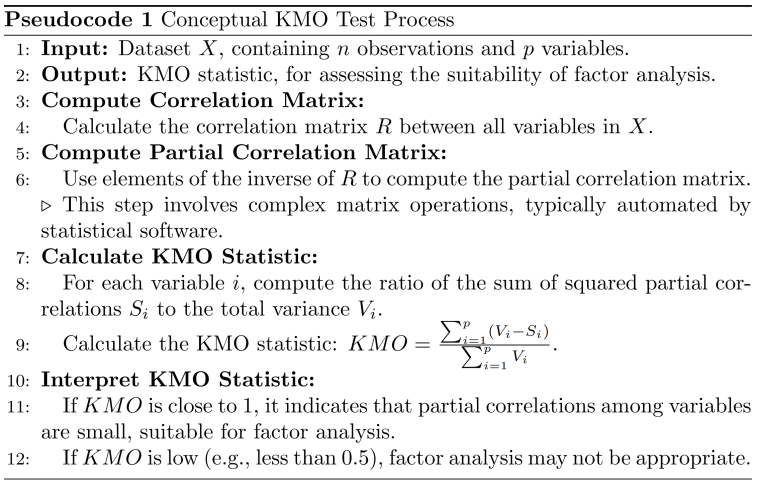}
\end{figure}

The magnitude of the KMO value shows a high degree of shared components among the variables, indicating that the dataset is suitable for factor analysis. Principal Component Analysis (PCA) takes a mathematical dimensionality reduction approach to find a few composite variables to replace the original multitude of variables, so that these composite variables can represent as much information as possible about the original variables and are uncorrelated with each other \cite{ref12}. The PCA algorithm can be found in Pseudocode 2.

\begin{figure}[h]
	\centering
	\includegraphics[width=0.46\textwidth]{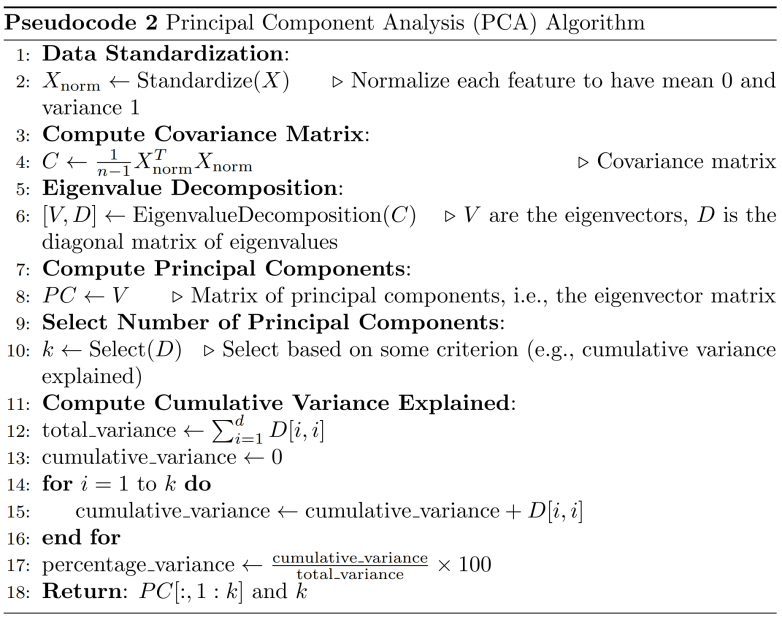}
\end{figure}

\begin{figure}[h]
	\centering
	\includegraphics[width=0.5\textwidth]{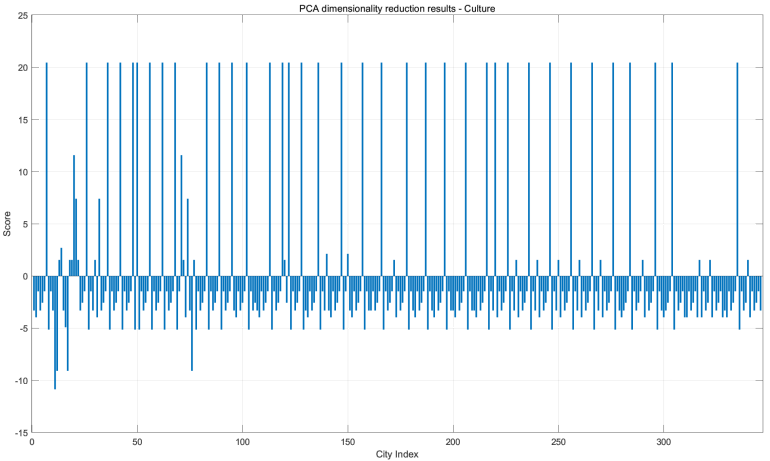}
	\includegraphics[width=0.5\textwidth]{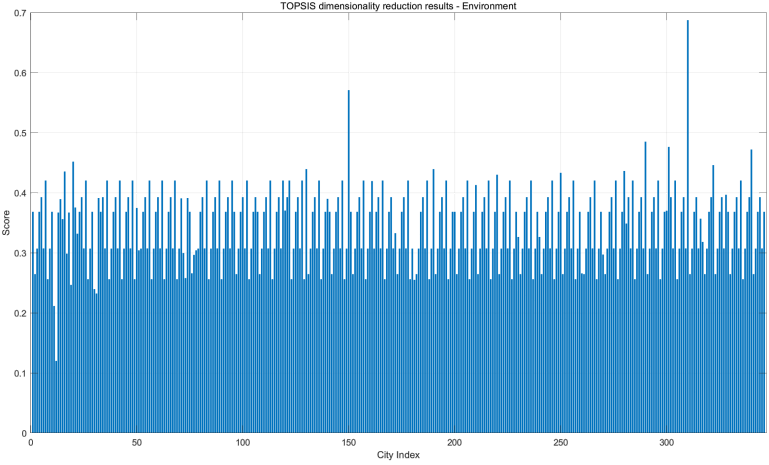}
	\caption{Dimensionality Reduction Results: MindSpore TOPSIS \& KMO}
	\label{fig:topsis_kmo}
\end{figure}

PCA can effectively reduce the dimensionality of variables while retaining the main variation information of the data, and is a commonly used technique for dimensionality reduction. The results of the TOPSIS and KMO downscaling are shown in Figure \ref{fig:topsis_kmo}.

The MindSpore framework was employed to conduct an in-depth analysis of a number of factors, including city size, ecological environment, depth of culture and history, accessibility, climatic characteristics, and specialties and cuisines. This was achieved through the utilisation of two advanced evaluation tools, namely the entropy weighting method and the TOPSIS evaluation method. Following a comprehensive evaluation process, 50 cities were identified as being particularly appealing to international travellers. The selection not only reflects the comprehensive strength of these cities, but also demonstrates their unique charms and characteristics. TOPSIS is a commonly used and effective multi-attribute decision-making method, which calculates the relative distances of each option from the theoretical optimal point and the theoretical disadvantage point to rank the options in order of merit. The TOPSIS algorithm is shown in Pseudocode 3.

\begin{figure}[h]
	\centering
	\includegraphics[width=0.48\textwidth]{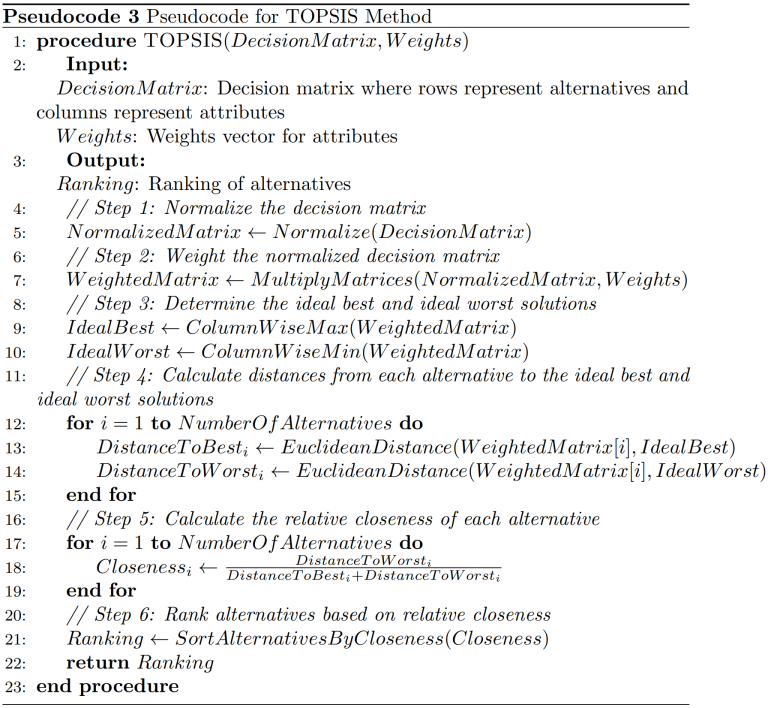}
\end{figure}

The final results of the MindSpore framework experiment are presented in the following Figure \ref{fig:topsis_score}. As can be observed from the figure, the city with the highest score is approximately 0.75, and the scores of the subsequent cities are more closely aligned, indicating that the overall indicators of these cities are relatively similar.

\begin{figure}[h]
	\centering
	\includegraphics[width=0.5\textwidth]{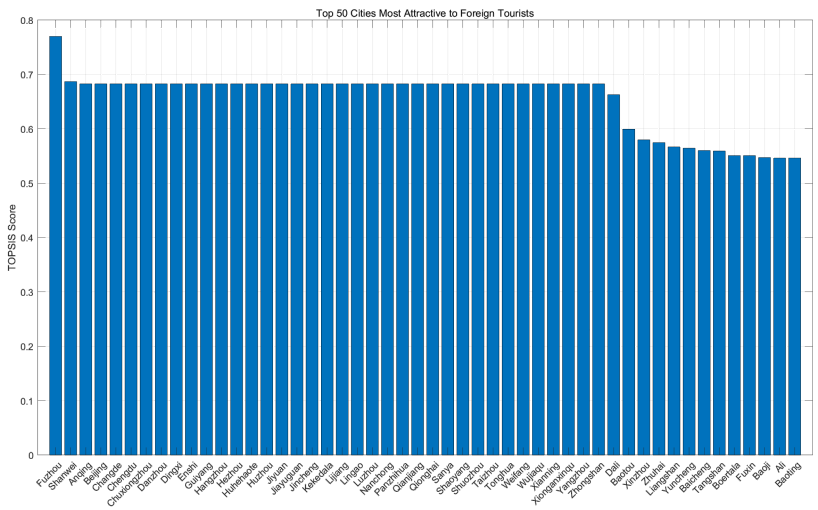}
	\caption{MindSpore TOPSIS Score}
	\label{fig:topsis_score}
\end{figure}

We build optimization models by calculating travel distance and time. High-Speed Rail Distance Calculation: The Haversine formula measures the spatial straight-line distance between any two points in a geographic coordinate system. The formula is specifically used to determine the shortest route distance between two locations on a spherical surface.\\

$
\resizebox{\columnwidth}{!}{$
	d = 2 \times 6371 \times \arcsin\left(\sqrt{\sin^2\left(\frac{\vartriangle\phi}{2}\right) + \cos(\phi_1) \times \cos(\phi_2) \times \sin^2\left(\frac{\vartriangle\lambda}{2}\right)}\right)
	$}
$\\

We made deeper extrapolations and extensions based on the preparatory work. For the part of city and latitude/longitude data: it involves the details of geographic coordinates of 352 cities so that we can accurately calculate the distance between different cities. For the mountain view information data: we compiled information about the most popular mountain views in each city, including the name of the view, its rating, the recommended time to visit, and information about the entrance fee.

\begin{enumerate}
	\item \textbf{Data Processing and Initialization:} Read and process the city and mountain view data to determine the starting parameters, which involves such aspects as total time constraints, time and cost at this stage, the city where it is located, and a list of cities and landscapes that have been previously traveled.
	\item \textbf{Selection of entry cities:} the first town to be reached is selected based on the highest rated mountain scenery.
	\item \textbf{Route planning:} A greedy algorithm is used to determine the next city to visit. All the cities that have not yet been visited are examined one by one, the distance from the high-speed train to the target city, the travel time and the cost are calculated, and finally the best rated mountain scenery with the lowest ticket price is selected for the tour. Calculate the statistics of the complete travel time, covering the time needed to travel by high-speed rail, move around locally, and visit tourist attractions, and incorporate rest periods. Prioritize the city with the lowest total cost as the next target to visit within the specified time.
	\item \textbf{Update current status:} Refresh the current city, date, expenses, list of visited cities and list of sightseeing landmarks. Repeat step 3 if you run out of time or are unable to locate the desired next destination.
	\item \textbf{Output Results:} Output the complete travel route, visited attractions, total travel time, total cost and total number of attractions.
\end{enumerate}

The greedy algorithm can find a solution that is close to the global optimum for the tourism route optimization problem in this study. Moreover, due to the selection of local optimal solutions at each step, the greedy algorithm has high execution efficiency and fast computation speed. The specific process of the greedy algorithm is shown in Pseudocode 4.

\begin{figure}[h]
	\centering
	\includegraphics[width=0.5\textwidth]{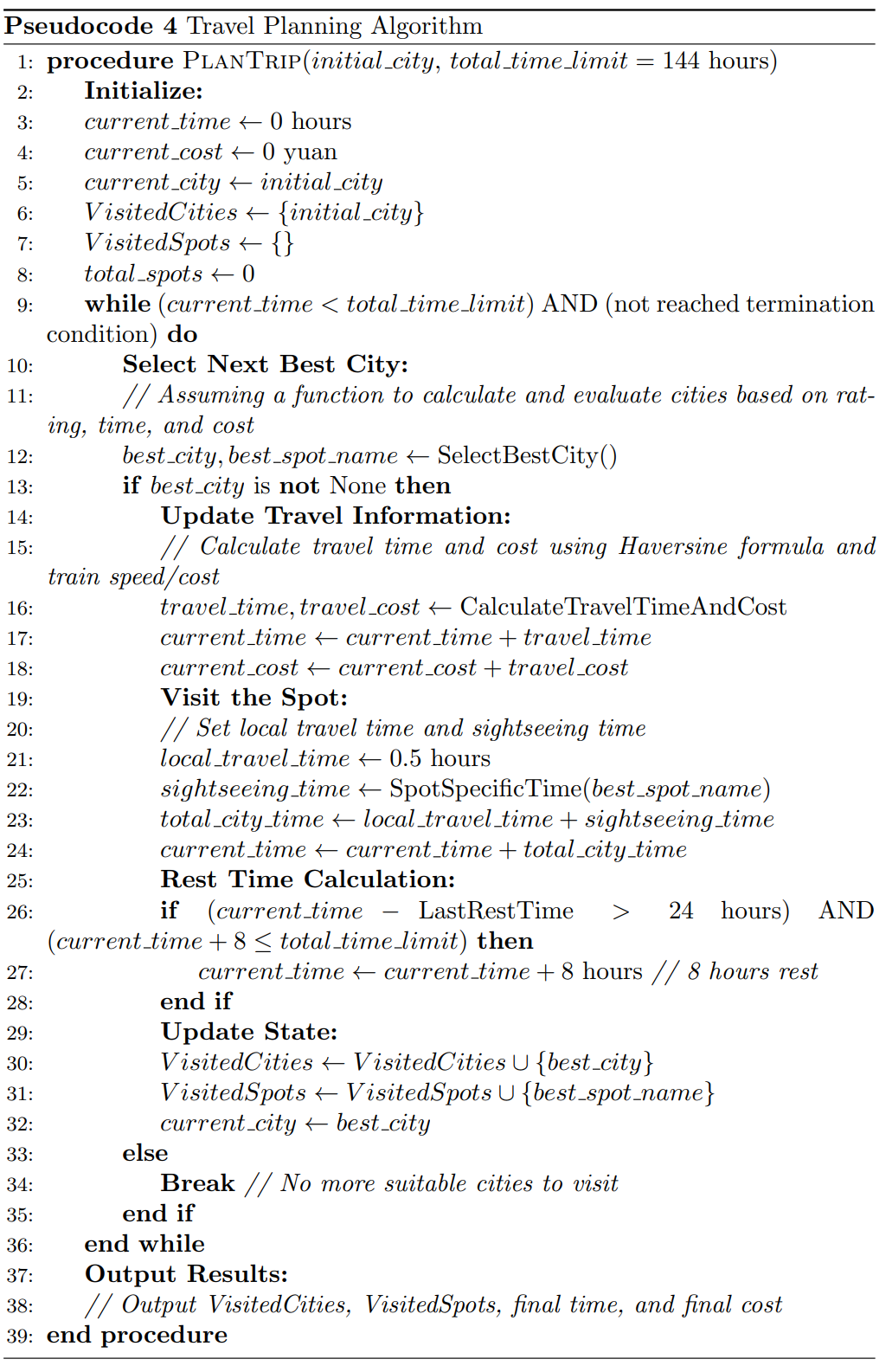}
\end{figure}

\begin{figure}[h]
	\centering
	\includegraphics[width=0.52\textwidth]{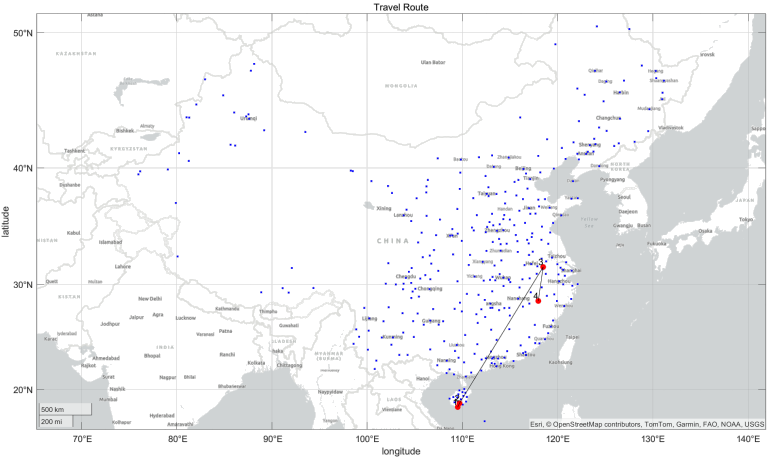}
	\caption{Travel Route}
	\label{fig:travel_route}
\end{figure}

At last, the travel planning routes can be generated using the greedy algorithm model within the MindSpore framework (see Figure \ref{fig:travel_route}). When devising a travel plan, factors such as tourists' interests, tour duration and budget will be considered in order to provide the most cost-effective travel plan.

Through the greedy algorithm, it is possible to combine scenic spot rating analysis, urban comprehensive evaluation, route optimization, cost calculation, and tourist preferences to output optimized tourism routes, ensuring the optimization and efficiency of the overall tourism planning scheme.

\section{Conclusions}

The findings of this study not only offer scientific and rational tourism planning solutions for international visitors but also illustrate the significant potential of greedy algorithms in the domain of path optimisation. In the future, the model will be further optimised by introducing the Octopus Algorithm \cite{ref13} or drawing on relevant cases of machine learning\cite{ref15, ref17, ref18, ref16, ref13, ref14}. This will enable the model to meet the personalised needs of different tourists, to keep up with the changing trends of the tourism market, and to contribute to the sustainable and high-quality development of the tourism industry. At the same time, the research result will be promoted to a wider range of tourism demand areas, thereby improving the tourism experience for a greater number of tourists.

\section{Acknowledgment}
Thanks for the support provided by MindSpore Community.

\printbibliography

\section*{Declaration of code and tool availability:}
Our results can be obtained through experiments using the MindSpore framework. The code for this article is available at    https://github.com/mindspore-lab/models/tree/master/research/arxiv\_papers
\end{document}